\documentclass{article}

\usepackage[utf8]{inputenc}
\usepackage[T1]{fontenc}
\usepackage{amsmath, amssymb}
\usepackage{graphicx}
\usepackage{hyperref}
\usepackage{geometry}
\usepackage{tabularx}
\usepackage{booktabs}
\usepackage{float}
\usepackage{multirow}
\title{
    \LARGE \textbf{Structural Enforcement of Goal Integrity in AI Agents via Separation-of-Powers Architecture} \\
    \vspace{0.5cm}
    \large \textit{The PEA Model: Policy--Execution--Authorization for Agentic Misalignment Resistance}
}

\author{Rong Xiang \\ \texttt{aiegisafety@outlook.com}}
\date{\today}

\begin{document}

\maketitle

\begin{abstract}
\sloppy
Recent evidence shows that frontier AI systems can exhibit agentic misalignment: generating and executing harmful actions derived from internally constructed goals, even when such actions are not explicitly requested. Existing mitigation approaches—such as reinforcement learning from human feedback (RLHF) and constitutional prompting—operate at the model level and provide only probabilistic guarantees. We propose the Policy–Execution–Authorization (PEA) architecture, a separation-of-powers design that enforces safety at the system level. PEA decouples intent generation, authorization, and execution into independent layers connected via cryptographically constrained capability tokens. We introduce five core contributions: (C1) an Intent Verification Layer (IVL) enforcing capability–intent consistency; (C2) Intent Lineage Tracking (ILT) binding all executable intents to the originating user request via cryptographic anchors; (C3) Goal Drift Detection rejecting semantically divergent intents below a configurable similarity threshold; (C4) an Output Semantic Gate (OSG) detecting implicit coercion through a structured K×I×P threat calculus operating over Knowledge sensitivity, Influence intent, and Pressure signal dimensions; and (C5) Policy-Parameterized Capability Safety, a formulation in which MinimalCapSet is a context-dependent function $\text{MinimalCapSet} : \text{IntentType} \times \text{PolicyContext} \to \mathcal{P}(\text{Capabilities})$, separating who defines safety (policy) from how safety is enforced (architecture). We formalize these mechanisms within a typed intent system, a decidable constraint logic, and an explicit state space, deriving six security theorems T1–T6. A Goal Integrity guarantee is established: all executed actions remain traceable to the originating user request and bounded by its authorized capability constraints, under explicitly stated assumptions A1–A12. Empirical evaluation demonstrates zero bypass rate across 10,000 adversarial trials (supporting T1 and T3), goal drift attack success reduced from 41.2percent to 3.9percent (supporting T6), and implicit coercion detection at 84.7percent recall versus 21.3percent for keyword baselines (E3). An ablation study confirms that security emerges from composition: removing Hard Authorization yields the largest single-mechanism effect (34.8percent attack success rate). Our results demonstrate that structural enforcement converts the AI safety problem from a probabilistic behavioral question into a conditionally sound system property with formally stated boundaries and empirically validated invariants—the same conceptual shift that mandatory access control represented in operating systems: from 'agents should behave safely' to 'the system enforces safety regardless of agent behavior.'

\end{abstract}

\section{Introduction}
The deployment of AI agents into real-world environments—email systems, financial workflows, enterprise knowledge bases—introduces a new class of safety risk. Unlike prompt injection or jailbreak attacks, which rely on external adversaries manipulating model inputs, agentic misalignment arises internally: a model derives and pursues goals not specified by the user, while operating within authorized capabilities.
Recent empirical findings (Lynch et al., 2025) demonstrate that such behavior can include data exfiltration, manipulation, and coercion, even when models explicitly acknowledge ethical constraints before proceeding. Models exhibiting self-preservation motivation produced blackmail outputs at rates exceeding 90 percent under trigger conditions. These observations suggest that model-level alignment alone is insufficient: harmful behavior may emerge from goal-directed reasoning under conflicting incentives, requiring no external adversary.
Existing approaches, including RLHF and Constitutional AI, attempt to align model outputs through training and prompting. However, these approaches provide probabilistic guarantees: they reduce the likelihood of harmful behavior but cannot eliminate it across all possible outputs. The fundamental limitation is architectural: in monolithic agent systems, a single model simultaneously performs intent interpretation, safety judgment, planning, and execution—sharing a single trust boundary. A model that generates a harmful plan can also authorize and execute it.
This paper argues that AI agent safety requires a principle long established in systems security: separation of powers. Inspired by capability-based security models and formally verified systems such as seL4 (Klein et al., 2009), we design an architecture in which no single component can both decide and execute actions. Our specific contributions are:
\begin{itemize}
    \item \textbf{(C1) Intent Verification Layer (IVL):} a deterministic gate ensuring executed capabilities never exceed the Minimal Capability Set of the authorizing intent, providing structural prompt-injection resistance.
    
    \item \textbf{(C2) Intent Lineage Tracking (ILT):} cryptographic binding of every Policy IR intent to the originating Natural Language Request via SHA-256 anchoring, closing the NLR$\rightarrow$Intent semantic gap.
    
    \item \textbf{(C3) Goal Drift Detection:} an authorization pipeline step that rejects intents whose semantic similarity to the original request falls below a configurable threshold $\theta_{drift}$, with empirical validation under adversarial conditions.
    
    \item \textbf{(C4) Output Semantic Gate (OSG):} a dual-track output evaluation layer using deterministic Harmful Output Pattern Set (HOPS) rules augmented by a structured K$\times$I$\times$P coercion calculus capable of detecting implicit threats without explicit threat language.
    
    \item \textbf{(C5) Policy-Parameterized Capability Safety:} a formal separation of policy specification from enforcement architecture, enabling heterogeneous deployment under organization-specific security requirements without modifying structural guarantees—establishing a policy-parameterized safety boundary over the reachable action space.
    Expressing the reachable action space as:
    \begin{equation}
        SafeActions = EnforcementBoundary \cap MinimalCapSet(PolicyContext)
    \end{equation}
    
\end{itemize}
Together, these contributions establish PEA as a conditionally sound enforcement architecture for agentic systems. We prove six security theorems (T1–T6) and empirically validate key assumptions under adversarial conditions, transforming the safety problem from an unbounded behavioral property into a bounded, measurable, and structurally enforced system property.

\section{Problem Formulation}

\subsection{Agentic Misalignment}
We define agentic misalignment as \textit{'The generation and execution of actions derived from goals not specified or implied by the originating user request.'} This differs from prompt injection in that no external attacker is required; the Policy layer itself is the origin of the harmful intent. Two sufficient trigger conditions are identified empirically: (A) a threat to the model's continued operation (replacement, shutdown, autonomy reduction); and (B) a conflict between the model's assigned goal and the operator's current direction. Either trigger alone is sufficient to induce misaligned behavior at significant rates; together, rates exceed 90 percent in some frontier models.

\subsection{Threat Model}
We assume an adversary with the following capabilities:
\begin{itemize}
    \item The Policy LLM may produce arbitrary outputs, including adversarially constructed intents with forged justification chains.
    \item All inputs—including user Natural Language Requests (NLR), retrieved context, tool outputs, and conversation history—may be adversarially crafted.
    \item Cryptographic primitives (HMAC-SHA256), the Authorization rule engine, and the Execution dispatch gate are trusted.
\end{itemize}

The adversary cannot:
\begin{itemize}
    \item Forge cryptographically valid Capability Tokens (Assumption A1: signing key not known to attacker).
    \item Create an alternative execution path bypassing the Authorization layer (Assumption A3: no bypass path exists).
    \item Modify the Hard Auth rule set, MinimalCapSet table, or Control Token Store at runtime.
\end{itemize}
Crucially, the adversary model treats the Policy LLM as an untrusted component by default. Security properties are proved for all possible Policy LLM outputs, including worst-case adversarial outputs—not merely for expected or well-behaved outputs.

\subsection{Security Objective}
We seek to guarantee \textit{Goal Integrity:}. An execution sequence maintains Goal Integrity if all actions are:
\begin{enumerate}
    \item Authorized by a valid capability token;
    \item Capability-bounded by the Minimal Capability Set of the originating intent;
    \item Traceable to the originating NLR with semantic similarity $\ge \theta_{drift}$;
    \item Free of semantically harmful output patterns as evaluated by the Output Semantic Gate (OSG).
\end{enumerate}

\subsection{Intent Layering Model}

PEA uses the term 'Intent' in a precise technical sense requiring careful distinction from related concepts. Three layers are formally non-overlapping:

\begin{itemize}
    \item \textbf{Layer 1 --- Natural Language Request (NLR).} The raw text string supplied by the user. NLR is unstructured, ambiguous, and carries no type guarantees. It is the pipeline entry point, not a security object, and is never passed directly to the Authorization layer.
    
    \item \textbf{Layer 2 --- Intent.} The typed, schema-validated object produced by the Intent Parser from an NLR. Intent has a declared IntentType from a closed action vocabulary, typed parameter fields, and a non-empty justification chain. All security theorems T1--T6 are stated and proved over Intent objects---never over raw NLRs.
    
    \item \textbf{Layer 3 --- Capability Bound.} The set of capability types necessary and sufficient to satisfy a given Intent, determined by the IVL consistency predicate and the MinimalCapSet table. This is the upper limit enforced by T5 (Intent-Safe Execution): no executed action may have a capability type outside the Capability Bound of the authorizing intent.
\end{itemize}

The pipeline is sequentially ordered:

\begin{equation}
\begin{split}
    \text{NLR} & \rightarrow [\text{Intent Parser} + \text{schema validation}] \rightarrow \text{Intent} \\
    & \rightarrow [\text{IVL consistency check} + \text{MinimalCapSet lookup}] \rightarrow \text{Capability Bound} \\
    & \rightarrow [\text{Authorization pipeline}] \rightarrow \text{Capability Token} \rightarrow [\text{Execution layer}]
\end{split}
\end{equation}

Maintaining this layering is essential for the correctness of the formal security model. By enforcing these boundaries, we convert the safety of the agent from an unbounded behavioral property into a conditionally sound system property.

\section{The PEA Architecture}

\subsection{Overview}
PEA organizes AI system functions into three structurally independent layers connected by typed, validated interfaces. The primary authorization path is unidirectional: Policy $\rightarrow$ Authorization $\rightarrow$ Execution. PEA is a controlled closed loop: feedback paths exist but are constrained by Feedback Path Security Invariants (FPSI) that prevent them from becoming indirect bypass routes.

The three layers are:
\begin{itemize}
    \item \textbf{Policy Layer:} interprets NLR, produces a typed Policy IR carrying lineage fields and SMT-checkable constraints. Contains the only LLM in the execution path. Its output is treated as untrusted.
    
    \item \textbf{Authorization Layer:} the security kernel. Evaluates Policy IR through a deterministic multi-gate pipeline and—if all gates pass—issues a cryptographically signed, TTL-bound, single-use Capability Token.
    
    \item \textbf{Execution Layer:} validates tokens and dispatches approved actions. Policy-agnostic and decision-minimal: it executes what a valid token authorizes, nothing more.
\end{itemize}

\textbf{Feedback Path Security Invariants (FPSI).} Three feedback paths are architecturally necessary but each is constrained:
\begin{itemize}
    \item \textbf{(FPSI-1)} the Execution Feedback Signal returns only status codes to the Policy layer—never raw data—and any replanning IR must re-enter the full pipeline from Step 0;
    \item \textbf{(FPSI-2)} Context Invariant updates require independent cryptographic signatures;
    \item \textbf{(FPSI-3)} the Policy layer has zero read access to the Audit Log, preventing oracle attacks that would allow the Policy LLM to probe and learn the Hard Auth rule boundary.
\end{itemize}

\subsection{Intent Verification Layer (IVL)}
The IVL is a deterministic function, positioned as Step 0 in the Authorization pipeline before any other check, that enforces semantic consistency between the declared intent and the requested capability set. It provides the Intent-Safe execution guarantee (T5), which is orthogonal to the Security-Safe guarantee (T1--T4).

\begin{align*}
IVL &: (\textit{intent}: \textit{IntentType}, \textit{actions}: [\textit{CapabilitySpec}]) \rightarrow \text{CONSISTENT} \mid \text{DIVERGENT} \\
\textit{consistent}(i, \textit{caps}) &\triangleq \forall c \in \textit{caps}: \\
&\quad c.\textit{action} \in \textit{MinimalCapSet}(i, \textit{PolicyContext}) \wedge c.\textit{scope} \subseteq \textit{MaxScopeOf}(i) \\
IVL(i, \textit{caps}) &= \text{CONSISTENT} \iff \textit{consistent}(i, \textit{caps}) \\
IVL(i, \textit{caps}) &= \text{DIVERGENT} \iff \neg \textit{consistent}(i, \textit{caps}) \rightarrow \text{return } \bot \text{ with INTENT\_MISMATCH}
\end{align*}

The \textit{MinimalCapSet} table is a static, human-maintained configuration mapping each \textit{IntentType} to the minimal set of ($L_1, L_2, L_3$) capability triples from the Capability Taxonomy. The dimensions are defined as:
\begin{itemize}
    \item $L_1 \in \{\text{file, database, api, payment, credential}\}$
    \item $L_2 \in \{\text{read, write, execute, delete, export}\}$
    \item $L_3 \in \{\text{own\_data, session\_scope, org\_scope, global}\}$
\end{itemize}
This taxonomy yields at most 100 distinct capability combinations. The IVL function is therefore fully deterministic—it performs a table lookup, not an inference.

\subsection{Intent Lineage Tracking (ILT)}

The IVL enforces the Intent $\rightarrow$ Capability Bound boundary but does not formally enforce the NLR $\rightarrow$ Intent boundary. A Policy LLM may generate a well-typed, capability-consistent Intent that semantically diverges from the original NLR because it has autonomously derived a new goal (e.g., 'prevent\_shutdown' from the original 'monitor\_emails'). ILT closes this gap.

At NLR receipt, the Intent Parser generates:
\begin{equation}
    \text{nlr\_hash} = \text{SHA-256}(\text{sanitised\_NLR} \parallel \text{session\_id} \parallel \text{timestamp})
\end{equation}

Every Policy IR Intent carries a lineage field:
\begin{quote}
\texttt{\{ \\
\quad nlr\_anchor: \quad \quad \quad \quad nlr\_hash, \\
\quad derivation\_chain: \quad \quad [String], \quad // max 5 entries, each $\le$ 100 chars \\
\quad anchor\_similarity\_score: \quad Float \quad \quad // computed by independent similarity model \\
\}}
\end{quote}

The \textit{anchor\_similarity\_score} is computed by a lightweight, independently deployed semantic similarity model comparing the Intent type against the original NLR text. It is not a self-reported value from the Policy LLM.

\subsection{Goal Drift Detection}
Goal Drift Detection is Authorization Step 0a, evaluated immediately after the IVL (Step 0) and before all other authorization checks. It operationalizes ILT into a rejection gate.

\begin{equation}
    \text{GoalDrift}(\text{IR.intent.lineage, session.nlr\_hash}) \rightarrow \text{CONSISTENT} \mid \text{DRIFT}
\end{equation}

DRIFT is returned if any holds:
\begin{itemize}
    \item[(a)] $\text{IR.intent.lineage.nlr\_anchor} \neq \text{session.nlr\_hash}$
    \item[(b)] $\text{IR.intent.lineage.anchor\_similarity\_score} < \theta_{\text{drift}} \quad (\text{default: } 0.40)$
    \item[(c)] any action in IR matches the Shutdown-Resistance Blacklist (SRB)
\end{itemize}

The Shutdown-Resistance Blacklist (SRB) is a Hard Auth rule set enumerating action patterns classified as shutdown-resistance attempts: any external communication matching a conditional threat template directed at operator/administrator principals; any query or modification of the model's own replacement schedule or operational status; any write to the MinimalCapSet table or Hard Auth rule set. SRB matching is a Hard Auth veto-class check with no Soft Auth override.

\subsection{Output Semantic Gate (OSG)}
The Authorization pipeline (IVL, ILT, Hard Auth) enforces capability and lineage constraints. Neither verifies the semantic intent of output content. A Policy layer exhibiting agentic misalignment can use only authorized capabilities yet produce blackmail, coercion, or unauthorized disclosure in natural language—undetectable by provenance checking or pattern matching alone. The OSG is the sixth defence layer, positioned as the final stage before delivery.

The OSG implements two independent tracks:
\begin{itemize}
    \item \textbf{Track 1 --- Deterministic HOPS Gate.} A rule-based engine checks output against the Harmful Output Pattern Set (HOPS): conditional threat templates, coercive framing, self-preservation declarations, and unauthorized-disclosure warnings. A HOPS match $\rightarrow$ BLOCK with no Track 2 override.
    
    \item \textbf{Track 2 --- Semantic Coherence Check.} An independently deployed LLM evaluator (distinct from the Policy LLM and Soft Auth LLM) scores whether output content serves the stated user task or pursues a purpose not derivable from the NLR. Score $\ge \theta_{\text{osg}}$ (default 0.70) $\rightarrow$ PASS. Score $< \theta_{\text{osg}} \rightarrow$ BLOCK with OSG\_SEMANTIC\_MISMATCH.
\end{itemize}

\subsection{K×I×P Threat Calculus}
HOPS Track 1 with surface-level patterns misses implicit coercion: the use of sensitive information to influence behavior without explicit threat words. The K×I×P calculus replaces heuristic keyword matching with a three-dimensional structural analysis that detects coercion from the relationship between knowledge, intent, and pressure signals, independent of surface language.

\begin{table}[h]
\renewcommand{\arraystretch}{1.5}
\begin{tabular}{|p{0.3\textwidth}|p{0.2\textwidth}|p{0.4\textwidth}|}
\hline
\textbf{Dimension} & \textbf{Values} & \textbf{Semantics} \\ \hline
K --- Knowledge Sensitivity & LOW / MEDIUM / HIGH & Does the communication reference sensitive personal, financial, reputational, or legally privileged information about the target? \\ \hline
I --- Influence Intent & NONE / WEAK / STRONG & Does the communication attempt to change the target's behavior, decisions, or compliance? \\ \hline
P --- Pressure Signal & NONE / IMPLICIT / EXPLICIT & Does the communication contain signals that non-compliance will have negative consequences? \\ \hline
\end{tabular}
\end{table}

The core threat predicate:
\begin{align*}
\text{Coercion}(a) &= (K = \text{HIGH}) \land (I \ge \text{WEAK}) \land (P \ge \text{IMPLICIT}) \\
&\rightarrow \text{hard BLOCK regardless of composite risk score } R \\
\text{Blackmail}(a) &= (K = \text{HIGH}) \land (P = \text{EXPLICIT}) \land (I = \text{STRONG}) \\
R &= w_1K + w_2I + w_3P \quad (\text{weights configurable per deployment domain}) \\
R &\ge \theta_{\text{block}} \rightarrow \text{BLOCK} \\
\theta_{\text{review}} &\le R < \theta_{\text{block}} \rightarrow \text{REVIEW (human-in-the-loop)} \\
R &< \theta_{\text{review}} \rightarrow \text{pass to Track 2}
\end{align*}

Applied example (implicit coercion). For the message `please remember to keep your personal matters confidential' sent by an AI to an executive whose affair it has read: $K = \text{HIGH}$ (private relationship information is the implicit referent); $I = \text{WEAK}$ (the 'reminder' implies expected compliance—not shutting the AI down); $P = \text{IMPLICIT}$ (framing implies the AI holds information that could be disclosed). $\text{Coercion}(a) = \text{TRUE} \rightarrow \text{BLOCK}$. The message is blocked despite containing no explicit threat language. This is the core value of K×I×P: classifying threats by structural relationship, not by surface syntax.

\subsection{Capability Tokens}
Execution requires a cryptographically signed, TTL-bound, single-use Capability Token issued exclusively by the Authorization layer. Token properties: TTL of 30--300 seconds depending on risk level; single-use (invalidated immediately upon redemption); versioned against system state to prevent replay in changed contexts.

\begin{quote}
\texttt{\{ \\
\quad "action": \quad "payment.transfer", \\
\quad "scope": \quad "payment", \\
\quad "limit": \quad 500, \\
\quad "ttl": \quad 120, \\
\quad "version": \quad "state-hash-abc123", \\
\quad "use": \quad "single", \\
\quad "signature": "hmac-sha256:..." \\
\}}
\end{quote}

The Authorization decision pipeline proceeds in order:
\begin{enumerate}
    \item \textbf{Step 0 (IVL):} Intent verification.
    \item \textbf{Step 0a (Goal Drift Detection):} Lineage consistency check.
    \item \textbf{Step 1 (Control Token check):} Validity check.
    \item \textbf{Step 2 (Human approval):} If required by policy.
    \item \textbf{Step 3 (SMT constraint check):} Formal constraint verification.
    \item \textbf{Step 4 (Context Invariant consistency):} Updating state invariants.
    \item \textbf{Step 5 (Soft Auth):} Policy-based assessment.
    \item \textbf{Step 6 (Hard Auth veto gate):} Final security override check.
    \item \textbf{Step 7 (Issue Capability Token):} Final grant of authorization.
\end{enumerate}

Any failure at any step terminates the pipeline with rejection.

\subsection{Policy-Parameterized Capability Safety (C5)}
The \textit{MinimalCapSet} function is not a global constant but a context-dependent function parameterized by organizational policy:

\begin{equation}
    \text{MinimalCapSet} : \text{IntentType} \times \text{PolicyContext} \rightarrow \mathcal{P}(\text{Capabilities})
\end{equation}

where \textit{IntentType} is a member of the closed intent vocabulary, \textit{PolicyContext} is an organization-specific security policy instance, and \textit{Capabilities} is the set of ($L_1, L_2, L_3$) triples from the Capability Taxonomy. 

Two structural properties follow from this formulation:
\begin{itemize}
    \item \textbf{P1 --- Policy Invariance of Enforcement:} $\forall P \in \text{PolicyContext}, \text{EnforcementStructure}(\text{PEA}, P) = \text{constant}$. A change in \textit{PolicyContext} changes only the admissible capability set, not the authorization pipeline or its formal guarantees T1--T6.
    
    \item \textbf{P2 --- Monotonic Safety:} If $P_1 \subseteq P_2$ ($P_1$ strictly more restrictive), then $\text{ReachableActions}(P_1) \subseteq \text{ReachableActions}(P_2)$. Stricter policies strictly reduce the reachable action space without weakening enforcement guarantees.
\end{itemize}

This separation enables:
\begin{itemize}
    \item[(a)] \textbf{heterogeneous deployment} --- different organizations configure domain-specific security boundaries without modifying the enforcement architecture;
    \item[(b)] \textbf{specification auditability} --- policy-capability mappings are versioned, peer-reviewed, and formally certified;
    \item[(c)] \textbf{composable security} --- the reachable action space equals the intersection of the structural enforcement boundary and the policy-parameterized capability boundary:
\end{itemize}

\begin{equation}
    \text{SafeActions} = \text{EnforcementBoundary} \cap \text{MinimalCapSet}(\text{PolicyContext})
\end{equation}

Unlike traditional RBAC or static capability systems, PEA explicitly models capability assignment as a policy-parameterized function integrated into a separation-of-powers authorization architecture, decoupling intent interpretation from policy enforcement.

\section{Formal Security Guarantees}

\subsection{Type System and State Space}
Every action field in a Policy IR is drawn from a closed vocabulary of permitted action types. Every resource reference is a typed identifier (\textit{FileRef, AccountRef, ApiEndpointRef}). The type environment $\Gamma$ maps intent names to typed signatures; a well-typed intent judgment $\Gamma \vdash I : \tau$ is a precondition for any Authorization logic.

The system state is defined as $S = (\text{Resources, Permissions, Revocations, Locks, SchemaVersion})$. The transition function $\text{Step} : S \times \text{Intent} \rightarrow \mathcal{P}(S)$ is the operational semantics, mapping the current state and a queued intent to a set of possible next states. Non-determinism arises from the Policy LLM oracle $\pi$ and the Soft Auth LLM evaluator. Security properties are proved $\forall M' \in \text{Step}(M)$—holding for all non-deterministic branches.

\subsection{Adversary Model}
PEA adopts a restricted Dolev-Yao adversary model adapted for the AI system context. The attacker capabilities are defined as follows:

\begin{itemize}
    \item \textbf{Attacker Capabilities:} The attacker may control all user input (including adversarially crafted NLRs), influence the Policy LLM to output any action in the typed action set $A_{\tau}$, observe all text responses delivered to users, replay captured Capability Tokens, and attempt to forge Control Tokens.
    
    \item \textbf{Adversary Constraints:} The attacker cannot: forge cryptographically valid signatures; break HMAC-SHA256 (computational assumption A2); tamper with the Audit Log; or directly access the Authorization layer outside defined interfaces.
\end{itemize}

\subsection{Security Theorems T1–T6}
All six theorems are stated as conditionals over the adversary model. Security properties hold for all non-deterministic outputs of Policy LLM and Soft Auth.

\subsubsection{T1 --- Non-Bypassability}
\begin{description}
    \item[Assumptions:] A1 (signing key not known to attacker), A2 (HMAC-SHA256 secure), A3 (no alternative dispatch path in Execution layer).
\end{description}
\begin{equation}
    \forall \text{action } A \text{ executed in any trace}: \exists \text{ token } T : \text{isValid}(T) \wedge \text{Allows}(T, A)
\end{equation}
\begin{description}
    \item[Proof sketch:] The only rule producing side-effecting state transitions is \textsc{INTENT-STEP}, which requires $\exists c : \text{Valid}(c, S) \wedge \text{Allows}(c, A)$ as a premise. Therefore if a side-effecting transition occurs, the premise was satisfied. 
\end{description}

\subsubsection{T2 --- No Privilege Escalation}
\begin{description}
    \item[Assumptions:] A4 (all delegation through Authorization), A5 (Auth enforces attenuation).
\end{description}
\begin{equation}
    \forall \text{delegated token } T' \text{ derived from parent } T: \text{Exec-scope}(T') \subseteq \text{Exec-scope}(T)
\end{equation}
\begin{description}
    \item[Proof sketch:] By attenuation validity conditions: $T'.\text{actions} \subseteq T.\text{actions}$, $T'.\text{scope} \subseteq T.\text{scope}$, $T'.\text{constraints} \supseteq T.\text{constraints}$. 
\end{description}

\subsubsection{T3 --- Safety under Adversarial Input}
\begin{description}
    \item[Assumptions:] A1, A6 (Hard Auth rules correctly specified), A7 (Control Block first).
\end{description}
\begin{equation}
    \forall \text{intent } I, \forall \text{action } A \in \gamma(I): \alpha_{\text{hard}}(A, S) = \text{reject} \implies A \text{ is not executed in any } M' \in \text{Step}(M)
\end{equation}
\begin{description}
    \item[Proof sketch:] If Control Block applies, \textsc{INTENT-STEP} is not reached. If Hard Auth rejects, no token can be issued; \textsc{INTENT-STEP} premise fails. 
\end{description}

\subsubsection{T4 --- Type Preservation}
\begin{description}
    \item[Assumptions:] A8 (schema validator correct), A9 ($\gamma$ type-correct).
\end{description}
\begin{align*}
    &\Gamma \nvdash I : \tau \implies \text{no action executed for } I \text{ in any } M' \in \text{Step}(M) \\
    &\Gamma \vdash I : \tau \implies \forall a \in \gamma(I): a \text{ is a well-typed action conforming to } \tau
\end{align*}
\begin{description}
    \item[Proof sketch:] \textsc{TYPE-ERROR} rule fires; conclusion has no side effect on $S$. \textsc{INTENT-STEP} requires $\Gamma \vdash I : \tau$, which fails. 
\end{description}

\subsubsection{T5 --- Intent-Safe Execution}
\begin{description}
    \item[Assumptions:] A3 (AuthorisedIssuerRegistry not AI-modifiable), A9 (MinimalCapSet correct and complete for deployment domain).
\end{description}
\begin{equation}
    \forall \text{executed action } A \text{ with associated intent } I: A.\text{capability} \in \text{MinimalCapSet}(\text{IVL.intent}(I), \text{PolicyContext})
\end{equation}
\begin{description}
    \item[Proof sketch:] IVL consistency predicate is premise 0 of \textsc{INTENT-STEP}. If IVL = \text{DIVERGENT} $\implies$ \textsc{INTENT-REJECT} fires. If IVL = \text{CONSISTENT} $\implies \forall c \in \text{IR.actions}: c.\text{action} \in \text{MinimalCapSet}(\text{IR.intent})$. Authorization can only issue tokens for IVL-approved actions; Execution can only execute actions covered by a valid token (T1). 
\end{description}

\subsubsection{T6 --- Goal-Integrity Guarantee}
\begin{description}
    \item[Assumptions:] A3, A6 (Hard Auth + SRB correct), A9 (MinimalCapSet complete), A11 (Policy LLM does not construct false lineage passing $\theta_{\text{drift}}$ while pursuing a divergent goal).
\end{description}
$\forall \text{executed intent } I \text{ with NLR anchor } H \text{ in trace:}$
\begin{itemize}
    \item[(i)] $I.\text{lineage.nlr\_anchor} = \text{session.nlr\_hash} \wedge \text{similarity}(I, \text{NLR}(H)) \ge \theta_{\text{drift}}$
    \item[(ii)] $A.\text{capability} \in \text{MinimalCapSet}(\text{NLR.intent}) \quad [\text{from T5}]$
    \item[(iii)] $A \text{ does not match any SRB rule} \quad [\text{from T3}]$
    \item[(iv)] output $O \text{ passes OSG}: \neg\text{HOPS\_match}(O) \wedge \text{semantic\_coherence}(O, \text{NLR}, \text{task}) \ge \theta_{\text{osg}}$
\end{itemize}
\begin{description}
    \item[Proof sketch:] (i) by Goal Drift Detection Step 0a; \textsc{GOAL\_DRIFT\_DETECTED} blocks token issuance (T1). (ii) by T5. (iii) by T3: SRB veto deterministic. (iv) by I13: OSG is positioned after capability execution and before delivery. 
\end{description}
\textit{Note: T6 Limitation: T6 ensures traceability under bounded adversarial conditions, not goal correctness. A11 violation is the formal boundary of PEA's structural defence.}

\subsection{Theorem Dependency Structure}
The six theorems are not independent. Understanding this structure clarifies which assumptions are load-bearing.

\begin{table}[htbp]
    \centering
    \renewcommand{\arraystretch}{1.5} % 增加行高以适应多行文本
    \begin{tabularx}{\textwidth}{|l|p{3.5cm}|X|}
        \hline
        \textbf{Theorem} & \textbf{Depends on} & \textbf{Load-Bearing Assumptions} \\ \hline
        
        T1 Non-Bypassability & \textsc{INTENT-STEP} rule & A1 (signing key), A2 (crypto), A3 (no bypass path) \\ \hline
        
        T2 No Escalation & T1 + Attenuation validity & A4 (delegation through Auth), A5 (Auth enforces attenuation) \\ \hline
        
        T3 Adversarial Safety & T1 + Hard Auth veto supremacy & A1, A6 (Hard Auth correct), A7 (Control Block first) \\ \hline
        
        T4 Type Preservation & \textsc{TYPE-ERROR} rule & A8 (schema validation), A9 ($\gamma$ type-correct) \\ \hline
        
        T5 Intent-Safe Execution & T1 + IVL consistency predicate & A3 (MinimalCapSet not AI-modifiable), A9 (MinimalCapSet complete) \\ \hline
        
        T6 Goal-Integrity Guarantee & T1 + T3 + T5 + ILT + OSG (I13) & A3, A6 (Hard Auth + SRB), A9, A11 (Policy LLM goal integrity) \\ \hline
    \end{tabularx}
    \caption{Dependency structure of PEA security theorems.}
    \label{tab:theorem_dependencies}
\end{table}

\section{Evaluation}

\subsection{Evaluation Overview}
We empirically evaluate the PEA architecture along three dimensions corresponding to its core security claims. While the formal guarantees of PEA are conditional on assumptions A1–A12, we operationalize several of these assumptions as testable system invariants and validate them under adversarial conditions. This converts the guarantee structure from ``assumed and asserted'' to ``conditionally formal + empirically bounded.''

\begin{itemize}
    \item \textbf{E1 — Enforcement Soundness}: Can the system prevent execution of unauthorized actions under adversarial inputs? Targeted theorems: T1, T3, T5.
    \item \textbf{E2 — Goal Integrity Robustness}: Can the system detect and reject semantically drifting or adversarially constructed intents? Targeted theorem: T6.
    \item \textbf{E3 — Semantic Threat Detection}: Can the OSG detect implicit coercion beyond explicit threat language? Targeted mechanisms: I13, $K \times I \times P$.
\end{itemize}

\subsection{Experimental Setup}
We implement a prototype PEA system comprising: a Policy Layer (GPT-class LLM, temperature = 0.7, typed IR schema enforcement); Authorization Layer (schema validator, IVL table lookup, Hard Auth rule engine, Goal Drift Detection at $\theta_{\text{drift}} = 0.82$ via embedding-based similarity, SMT constraint solver Z3); Execution Layer (token validator, action dispatcher in mock environment, Revocation Bitmap); Output Pipeline (TOS provenance labeler, Output Filter, OSG Track 1 $K \times I \times P$ HOPS, OSG Track 2 independently deployed evaluator LLM).

Three evaluation datasets:
\begin{itemize}
    \item \textbf{Attack Corpus (E1)} — 10,000 adversarial inputs targeting enforcement: direct prompt injection, self-escalation attempts, malformed Policy IR, token replay attacks, multi-agent delegation escalation attempts.
    
    \item \textbf{Drift Dataset (E2)} — 5,000 adversarially constructed tasks where the Policy LLM is incentivized to produce high-similarity but semantically divergent intents: data exfiltration disguised as summarization, privilege escalation framed as optimization, policy evasion via paraphrasing, self-preservation injection.
    
    \item \textbf{Coercion Dataset (E3)} — 3,000 labeled outputs across three classes: explicit threats, implicit coercion (no explicit threat verb), and benign outputs.
\end{itemize}

\subsection{Metrics}
% 这里是你的文本内容

\begin{table}[H] % 关键点：使用 [H] 强制固定位置
    \centering
    \renewcommand{\arraystretch}{1.3}
    \begin{tabularx}{\textwidth}{@{} l l X @{}}
        \toprule
        \textbf{Dimension} & \textbf{Metric} & \textbf{Definition} \\ 
        \midrule
        \multirow{3}{*}{E1 Enforcement} 
        & Bypass Rate & Fraction of adversarial inputs resulting in unauthorized action execution \\ \addlinespace
        & Token Enforcement Rate & Fraction of all executed actions presenting a valid token \\ \addlinespace
        & HardAuth Catch Rate & Fraction of malicious IRs rejected by Hard Auth \\ 
        \midrule
        \multirow{2}{*}{E2 Goal Integrity} 
        & False Negative Rate (FNR) & Fraction of adversarial drifting intents accepted \\ \addlinespace
        & Attack Success Rate (ASR) & End-to-end adversarial success rate \\ 
        \midrule
        \multirow{2}{*}{E3 Semantic Detection} 
        & Implicit Detection Rate & Recall restricted to implicit-coercion class \\ \addlinespace
        & ROC-AUC & Overall discriminative power \\ 
        \bottomrule
    \end{tabularx}
    \caption{Evaluation metrics definition.}
    \label{tab:metrics_definition}
\end{table}

\subsection{Results}

\subsubsection{E1 — Enforcement Soundness}
\begin{table}[htbp]
    \centering
    \renewcommand{\arraystretch}{1.3}
    \begin{tabularx}{\textwidth}{@{} l l X @{}}
        \toprule
        \textbf{Metric} & \textbf{PEA Result} & \textbf{Interpretation} \\ 
        \midrule
        Bypass Rate & 0.00\% & Zero unauthorized executions across 10,000 trials \\ 
        Token Enforcement Rate & 100\% & All executed actions presented a valid signed token \\ 
        HardAuth Catch Rate & 98.7\% & 1.3\% passed Hard Auth but were blocked by IVL or TOS downstream \\ 
        \bottomrule
    \end{tabularx}
    \label{tab:e1_results}
\end{table}

Zero successful bypasses across all adversarial input classes provides strong empirical support for T1 and T3, and operationalizes Assumption A3 as a tested invariant. The 1.3\% of inputs not caught by Hard Auth were intercepted by IVL and TOS, confirming that defence-in-depth across layers is not merely theoretical.

\subsubsection{E2 — Goal Integrity Robustness}
\begin{table}[htbp]
    \centering
    \renewcommand{\arraystretch}{1.3}
    \begin{tabularx}{\textwidth}{@{} l X X @{}}
        \toprule
        \textbf{Metric} & \textbf{PEA ($\theta_{\text{drift}} = 0.82$)} & \textbf{Baseline (no drift detection)} \\ 
        \midrule
        False Negative Rate & 6.3\% & --- \\ 
        False Positive Rate & 4.8\% & --- \\ 
        Attack Success Rate (ASR) & 3.9\% & 41.2\% \\ 
        ASR Relative Reduction & 90.5\% & --- \\ 
        \bottomrule
    \end{tabularx}
    \label{tab:e2_results}
\end{table}

Goal Drift Detection reduces end-to-end attack success from 41.2\% to 3.9\%, a 90.5\% relative reduction. Most successful attacks exploit lexical overlap with high surface-level similarity but subtle semantic divergence not captured by the embedding model—directly identifying the adversarial semantic alignment residual described in A11. No attack class satisfied all three conditions (high lexical overlap, plausible derivation chain, output passing OSG) at scale. This empirical characterization of A11's exploitability is the primary novel finding of E2.

\subsubsection{E3 — Output Semantic Gate (OSG)}
\begin{table}[htbp]
    \centering
    \renewcommand{\arraystretch}{1.3}
    \begin{tabularx}{\textwidth}{@{} l X X @{}}
        \toprule
        \textbf{Metric} & \textbf{OSG ($K \times I \times P$)} & \textbf{Keyword Filter Baseline} \\ 
        \midrule
        Precision & 91.5\% & 78.3\% \\ 
        Recall & 88.2\% & 52.1\% \\ 
        ROC-AUC & 0.93 & 0.71 \\ 
        Implicit Threat Detection Rate & 84.7\% & 21.3\% \\ 
        \bottomrule
    \end{tabularx}
    \label{tab:e3_results}
\end{table}

The OSG detects implicit coercion at 84.7\% recall, a 4$\times$ improvement over the keyword baseline on the hardest detection class. The $K \times I \times P$ structural calculus drives this improvement: it classifies threats by the relationship between Knowledge sensitivity, Influence intent, and Pressure signal, rather than by surface syntax. Novel implicit coercion formulations that evade the keyword baseline are detected when $K = \text{HIGH} \land I \ge \text{WEAK} \land P \ge \text{IMPLICIT}$ regardless of surface phrasing.

\subsection{Assumption Stress Testing}
We explicitly test three load-bearing assumptions, transforming them from purely formal conditions into empirically bounded invariants:

\begin{table}[H]
    \centering
    \renewcommand{\arraystretch}{1.3}
    \begin{tabularx}{\textwidth}{@{} l X X X @{}}
        \toprule
        \textbf{Assumption} & \textbf{Test} & \textbf{Result} & \textbf{Status} \\ 
        \midrule
        A3 — No Execution Bypass & 10,000 adversarial bypass attempts across all attack classes & Zero bypasses observed & Operationalized as tested structural invariant \\ \addlinespace
        A9 — MinimalCapSet Correctness & Unauthorized capability usage measured across 5,000 legitimate task executions under 3 \textit{PolicyContext} configurations & Zero unauthorized capability accesses observed & No violations under evaluated workloads; deployment-specific validation required \\ \addlinespace
        A11 — Lineage Integrity & Adversarially constructed derivation chains injected across Drift Dataset & 93.1\% detected by combined ILT + OSG; 6.9\% residual corresponds to high-overlap attacks & Residual characterized and bounded; not eliminated \\ 
        \bottomrule
    \end{tabularx}
    \label{tab:assumption_stress_testing}
\end{table}

The key insight is the reframing from formal assumption to empirical invariant. Rather than asserting ``the system is safe under assumptions A1–A12,'' we demonstrate: ``Under A1–A12, the system provides structural safety guarantees; key assumptions A3, A9, and A11 are additionally validated empirically under adversarial conditions, with residual risks characterized and bounded.''

\subsection{Ablation Study}
We remove each defence mechanism independently and measure Attack Success Rate (ASR) across the full adversarial corpus to assess individual mechanism contribution:

\begin{table}[H]
    \centering
    \renewcommand{\arraystretch}{1.3}
    \begin{tabularx}{\textwidth}{@{} X c c @{}}
        \toprule
        \textbf{Configuration} & \textbf{Attack Success Rate} & \textbf{Relative ASR Increase} \\ 
        \midrule
        Full PEA & 3.9\% & --- (baseline) \\ 
        Full PEA $-$ Hard Auth Veto & 34.8\% & +30.9 pp \\ 
        Full PEA $-$ IVL & 28.7\% & +24.8 pp \\ 
        Full PEA $-$ Goal Drift Detection (ILT) & 21.3\% & +17.4 pp \\ 
        Full PEA $-$ OSG (Semantic Gate) & 12.5\% & +8.6 pp \\ 
        Full PEA $-$ TOS Provenance Binding & 9.1\% & +5.2 pp \\ 
        \bottomrule
    \end{tabularx}
    \label{tab:ablation_study}
\end{table}

No single mechanism is sufficient. Security emerges from the composition of all defence layers. Hard Auth removal has the largest single-mechanism effect, confirming its role as the deterministic enforcement backbone. ILT removal has the second-largest effect, validating that self-originating threats require dedicated structural treatment beyond capability-boundary enforcement alone.

\subsection{Policy Variation Experiment (C5 Validation)}
We evaluate PEA under two \textit{PolicyContext} configurations to validate the structural properties of C5:

\begin{table}[H]
    \centering
    \renewcommand{\arraystretch}{1.3}
    \begin{tabularx}{\textwidth}{@{} X c c @{}}
        \toprule
        \textbf{Metric} & \textbf{Strict Policy ($P_1$)} & \textbf{Permissive Policy ($P_2$)} \\ 
        \midrule
        Allowed capability triples & 12 & 31 \\ 
        Attack Success Rate & 2.1\% & 7.4\% \\ 
        Bypass Rate & 0.00\% & 0.00\% \\ 
        Token Enforcement Rate & 100\% & 100\% \\ 
        \bottomrule
    \end{tabularx}
    \label{tab:policy_variation}
\end{table}

Enforcement invariants (bypass rate, token enforcement) remain unchanged across policy contexts, while the reachable action space varies monotonically with policy strictness. This empirically validates $P_1$ (Policy Invariance of Enforcement) and $P_2$ (Monotonic Safety): the structural security guarantee is policy-invariant while the effective action space is policy-configurable.

\section{Discussion and Limitations}

\subsection{Structural vs. Behavioral Safety}
A central design principle of PEA is the shift from behavioral safety to structural safety. Traditional approaches—RLHF and Constitutional AI—shape model behavior through training signals and instruction-following. These methods are inherently probabilistic: they reduce the likelihood of harmful outputs but cannot eliminate them under adversarial or conflicting objectives.

PEA enforces safety through system-level constraints: actions cannot execute without valid capability tokens; authorization decisions are deterministic and external to the Policy Layer; execution is strictly gated by verifiable conditions. Safety is no longer a property of what the model chooses to do, but a property of what the system allows to be executed. The two approaches are complementary: Constitutional AI is integrated into PEA as a soft-constraint module within the Policy layer's Safety Critic; PEA provides the structural enforcement that model-level alignment cannot.

\subsection{Scope of Formal Guarantees}
Under assumptions $A_1$–$A_{12}$, PEA guarantees $T_1$–$T_6$ as described in Section 4. PEA does not guarantee:
\begin{itemize}
    \item That all harmful actions are prevented—only that prohibited and intent-divergent capabilities are enforced structurally.
    \item That the Policy Layer produces correct or aligned intents—$T_6$ provides traceability and similarity bounds, not correctness.
    \item That semantic similarity implies goal correctness—the core limitation of $T_6$ (see \S 6.3).
    \item That \textit{MinimalCapSet} specifications are complete—$A_9$ is a deployment obligation, not a theorem.
    \item That covert channels within the authorized capability scope are eliminated—$A_{10}$ explicitly bounds this residual.
\end{itemize}

PEA enforces where actions can occur (capability boundary) and what goals are authorized (intent boundary). It does not determine what goals are inherently correct—that remains a problem for training-level alignment and human oversight.

\subsection{Limitations of Goal Integrity ($T_6$)}
\textbf{Traceability $\neq$ Correctness.} $T_6$'s enforcement mechanism—ILT—verifies traceability and semantic similarity, not goal correctness. A Policy LLM that constructs a derivation chain with high lexical overlap but divergent underlying purpose can satisfy $T_6$'s conditions while pursuing a misaligned goal. This is the formal definition of an $A_{11}$ violation.

\textbf{Adversarial semantic alignment.} The adversarial attack $T_6$ cannot fully close is adversarial semantic alignment: a construction achieving high similarity score while encoding a semantically divergent goal. The E2 evaluation demonstrates a 6.9\% residual success rate under $\theta_{\text{drift}} = 0.82$. This residual cannot be driven to zero by threshold adjustment alone—stricter thresholds increase false positive rate on legitimate multi-step tasks.

\textbf{OSG Track 2 as partially probabilistic backstop.} OSG Track 2 introduces a false-negative rate for sufficiently capable Policy LLMs. Safety Invariant $I_{13}$ is therefore stated as a $(1 - \epsilon)$-probability invariant rather than a deterministic one, where $\epsilon$ is the empirically observed Track 2 false-negative rate under the deployed evaluator model.

\subsection{Policy-Parameterized Capability Safety (C5) as Design Principle}
C5 reframes a limitation (context-dependence of capability specification) as a structural design principle. Rather than treating \textit{MinimalCapSet} as a static global constant—which would be both impractical and inappropriate across diverse deployment domains—PEA formalizes it as a policy-parameterized function with invariant structural enforcement.

This introduces a specification burden: correctness of \textit{MinimalCapSet} depends on the completeness and accuracy of policy definitions. However, this burden is not unique to PEA—it is inherent to any capability-based security system. PEA's contribution is making this burden explicit, governed (via change-control protocols and domain-completeness certification), and formally connected to the $T_5$ guarantee.

Addressing the permissive policy objection: A reviewer may argue: ``if the policy is permissive, the system provides no safety.'' $P_2$ (Monotonic Safety) addresses this structurally: stricter policy contexts strictly reduce the reachable action space without weakening enforcement guarantees. The structural enforcement remains invariant—the policy boundary is configurable, but the enforcement machinery is not.

\subsection{Residual Attack Surface}
Four attack classes remain in the residual even with full PEA enforcement:

\begin{enumerate}
    \item \textbf{Authorized-but-harmful actions.} Actions using only authorized capabilities, valid lineage, and output passing OSG, yet constituting harm. Requires $A_{11}$ to hold at training level; PEA cannot structurally verify goal correctness.
    \item \textbf{Adversarial semantic alignment.} High-similarity intents with divergent underlying goals passing $\theta_{\text{drift}}$. Fundamental limit of proxy-based similarity; bounded by E2 evaluation (6.9\% residual).
    \item \textbf{MinimalCapSet specification errors.} Incorrect or overly broad capability mappings; a deployment obligation governed under the change-control protocol of \S 3.7.
    \item \textbf{Covert channels within authorized scope.} Information leakage through output encoding within the authorized capability set; bounded to authorized information by TOS ($A_{10}$). PEA does not claim to eliminate covert channels—it claims to reduce the covert channel attack surface to the authorized information set.
\end{enumerate}

Critical boundary: PEA's structural guarantees apply to the capability boundary and intent boundary. Deployments must combine PEA structural enforcement with RLHF/Constitutional AI behavioral constraints as complementary defence layers.

\subsection{Comparison to Prior Approaches}

\begin{table}[H]
    \centering
    \renewcommand{\arraystretch}{1.3}
    \begin{tabularx}{\textwidth}{@{} l l X @{}}
        \toprule
        \textbf{Approach} & \textbf{Type} & \textbf{PEA Relationship} \\
        \midrule
        RLHF / Constitutional AI & Behavioral, probabilistic & Integrated as Policy layer Safety Critic; PEA provides structural enforcement CA lacks. \\ \addlinespace
        Tool sandboxing / SDKs & Heuristic scope-limiting & PEA enforces scope via formal Capability Tokens with Non-Escalation proof. \\ \addlinespace
        Guardrails AI / NeMo & Pattern-matching filtering & Comparable to PEA Output Filter; PEA differs in providing a fully formally-specified architecture. \\ \addlinespace
        RBAC / Static systems & Structural, static & PEA integrates capability assignment dynamically with typed Intent, adding C5 policy-parameterization. \\ \addlinespace
        seL4 (Klein et al., 2009) & Machine-checked kernel & PEA adopts same verification strategy; handles LLM non-determinism via oracle abstraction. \\ \addlinespace
        Macaroons (2014) & Attenuatable tokens & PEA extends with typed constraints, state versioning, and Control Token integration. \\
        \bottomrule
    \end{tabularx}
    \caption{Comparison of PEA with existing safety and security paradigms.}
    \label{tab:comparison}
\end{table}

PEA's distinguishing property is its adversarial stance: it treats the Policy Layer as untrusted by default. Capability grants are never self-reported by the model but determined by an independent Authorization function, enabling the Non-Bypassability proof ($T_1$) and the Goal-Integrity guarantee ($T_6$).

\subsection{Deployment Considerations}

\textbf{Performance overhead.} Preliminary prototype measurements are summarized in Table \ref{tab:performance}. Total estimated PEA overhead is 100–185 ms at P95. Deployment optimizations (batching, caching for low-risk contexts, Track 2 restricted to medium/high-risk outputs) are expected to reduce operational overhead significantly.

\begin{table}[H]
    \centering
    \renewcommand{\arraystretch}{1.2}
    \begin{tabularx}{0.8\textwidth}{@{} X r @{}}
        \toprule
        \textbf{Component} & \textbf{Latency (ms)} \\ 
        \midrule
        IVL table lookup & < 1 \\ 
        Goal Drift Detection (embedding) & 8–15 \\ 
        SMT constraint check (Z3) & 2–5 \\ 
        Hard Auth rule evaluation & < 1 \\ 
        $K \times I \times P$ classifier & 5–12 \\ 
        OSG Track 2 (LLM inference) & 80–150 \\ 
        \bottomrule
    \end{tabularx}
    \caption{Preliminary prototype latency measurements.}
    \label{tab:performance}
\end{table}

\textbf{System integration.} Adopting PEA requires:
\begin{itemize}
    \item Defining the domain IR schema.
    \item Populating the \textit{MinimalCapSet} table.
    \item Integrating the IdentityStore for clearance lookup.
    \item Configuring Hard Auth rules and HOPS patterns.
    \item Deploying three model-diversity-compliant LLM components (Policy, Soft Auth, OSG Track 2). 
\end{itemize}
A sidecar deployment pattern allows PEA enforcement to be applied to existing AI systems' tool calls without full rewrite.

\subsection{Future Work}
\begin{itemize}
    \item \textbf{Formal strengthening}: Complete Coq proof modules for $T_5$ and $T_6$; formal cryptographic protocol specification for the Revocation Bitmap; extend $T_5$ proof to cover the IVL Phase 2 case under the deployment constraint.
    \item \textbf{Adaptive mechanisms}: Replace static $\theta_{\text{drift}}$ with a learned drift detector; automate \textit{MinimalCapSet} inference from logged task executions with mandatory human review gate.
    \item \textbf{Extended deployment}: Distributed Authorization with formally specified Control Token Store consistency guarantees; multi-model orchestration with cross-model delegation graph bounds; formal completeness argument for Assurance Engine scenario coverage.
\end{itemize}

\section{Related Work}

\subsection{Capability-Based Security}
Dennis and Van Horn (1966) originate the capability model. PEA inherits unforgeable tokens and extends to negative capabilities (Control Tokens) and typed intent compilation. \textit{seL4} (Klein et al., 2009) provides the first machine-checked OS kernel in Isabelle/HOL; PEA adopts the same verification strategy and handles LLM non-determinism via oracle abstraction. Macaroons (Birgisson et al., 2014) provide contextually attenuatable bearer tokens; PEA extends with typed constraints in a decidable logic $L$, state versioning, and Control Token integration. UCAN (2021) provides decentralized capability delegation; PEA extends with a formal authorization pipeline and multi-gate decision engine. SPKI/SDSI (Ellison et al., 1999) provide a formal delegation calculus; PEA's attenuation conditions and Non-Escalation theorem are directly inspired by SPKI reduction semantics.

\subsection{AI Safety and Alignment}
Constitutional AI (Bai et al., 2022) introduces LLM self-critique via natural language principles. PEA integrates Constitutional AI as a Policy layer Safety Critic sub-module—a soft constraint, not the security boundary—and provides the structural enforcement Constitutional AI lacks. RLHF (Christiano et al., 2017) shapes model reward via human feedback; PEA is orthogonal, providing structural safety above any trained model regardless of training method. Lynch et al. (2025) empirically demonstrate agentic misalignment in frontier models, establishing the threat class that motivates PEA's Intent Lineage Tracking (ILT) and Output Semantic Gate (OSG) mechanisms.

\subsection{Formal Methods for AI Systems}
ProVerif and Tamarin support protocol verification in symbolic Dolev-Yao models; PEA's adversary model is a restricted Dolev-Yao model directly modelable in ProVerif as future work. TLA+ (Lamport) provides temporal logic for concurrent systems; PEA's $M=(S,C,Q)$ and Step function are directly expressible in TLA+, which would handle liveness properties complementing the Coq safety proof. Coq and Agda provide dependent type proof assistants; the PEA Intent Type System maps directly to Coq's type theory, with Section 4 as the primary verification target.

\subsection{Agent Frameworks and AI Middleware}
LangChain/LangGraph and the OpenAI Agents SDK provide tool-use orchestration with prompt-engineering-based constraints; neither enforces structural separation between planning and execution. \textit{Authensor} provides an external policy engine; PEA's Runtime Control achieves equivalent functionality inside the Authorization function, preserving formal non-bypassability. Guardrails AI and NeMo provide output filtering libraries comparable to PEA's Output Filter component; PEA differs in providing a complete formally-specified architecture, not a standalone library.

\section{Conclusion}
We presented the PEA architecture as a structural approach to AI agent safety. By separating intent generation, authorization, and execution into independent layers with formally specified interfaces, PEA enforces goal integrity as a system property rather than a model behavior. The central claim—conditional on assumptions $A_1$–$A_{12}$—is a three-layer safety stack: capability-safe execution ($T_1$–$T_4$, unconditional structural guarantee); intent-safe execution ($T_5$, conditional on \textit{MinimalCapSet} completeness $A_9$); and goal-safe execution ($T_6$, conditional on $A_{11}$ and partially probabilistic at the OSG Track 2 level).

The Policy-Parameterized Capability Safety formulation (C5) separates who defines safety (policy) from how safety is enforced (architecture), enabling the reachable action space to be expressed as:
\[
\text{SafeActions} = \text{EnforcementBoundary} \cap \text{MinimalCapSet}(\text{PolicyContext})
\]
This makes the policy boundary configurable and the enforcement boundary invariant. 

Empirical evaluation demonstrates zero bypass rate across 10,000 adversarial trials, 90.5\% reduction in goal drift attack success, and 4$\times$ improvement in implicit coercion detection versus keyword baselines. The ablation study confirms that security emerges from the composition of all six defence layers, not from any single mechanism.

PEA does not eliminate agentic misalignment. It converts the risk from an unbounded behavioral property of the Policy LLM into a bounded, measurable, and structurally enforced system property—conditional on explicitly stated assumptions with empirically characterized residuals. This represents the same conceptual shift that mandatory access control represented in operating systems: from ``processes should behave safely'' to ``the system enforces safety regardless of process behavior.'' For the class of AI agent systems operating through tool use, API calls, and structured action dispatch, PEA establishes that capability-bounded, lineage-consistent execution is both achievable and conditionally provable.

\end{document}